\documentclass[runningheads]{llncs}

\usepackage[T1]{fontenc}
\usepackage{graphicx}
\usepackage{amsmath,amssymb}
\usepackage{booktabs}
\usepackage{multirow}
\usepackage{xcolor}
\usepackage{nicefrac}
\usepackage{microtype}
\usepackage[hidelinks,breaklinks=true,bookmarks=false]{hyperref}

\newcommand{\etal}{\textit{et al.}}
\newcommand{\eg}{\textit{e.g.}}

\begin{document}

\title{When Two Tracers Disagree: An Investigation of Multimodal Fusion for Clinical PET/CT Segmentation}

\titlerunning{Multimodal Fusion for Clinical PET/CT Segmentation}
\author{Jack A. Johnson\inst{1,2} \and Bartłomiej W. Papież\inst{3}}
\authorrunning{J. A. Johnson and B. W. Papież}
\institute{Nuffield Department of Medicine, University of Oxford, UK \and
Department of Oncology, University of Oxford, UK \and
Big Data Institute, Nuffield Department of Population Health, University of Oxford, Oxford, United Kingdom}

\maketitle

\begin{abstract}
PSMA and FDG PET/CT visualise complementary biological information in prostate cancer.
Combining both tracers could capture heterogeneous tumour phenotypes that may be missed by either alone, yet there is no consensus on effective deep learning architectures for fusing these modalities.
We evaluated multimodal image-fusion strategies for automatic whole-body PET/CT lesion segmentation to estimate total tumour burden.
Using the public DEEP-PSMA Challenge dataset, we trained tracer-specific 3D nnU-Net baselines and compared (i)~early fusion with a single encoder and one decoder (OEOD) or two decoders (OETD), and (ii)~intermediate fusion via a dual-encoder cross-attention U-Net (DECA-UNet).
Tracer-specific baselines performed strongly (PSMA Dice\,=\,0.93; FDG\,=\,0.81).
Fusion yielded mixed results: OEOD produced a combined Dice of 0.90 (on an easier, non-tracer-specific task), whilst the tracer-specific fusion models reached PSMA/FDG\,=\,0.69/0.64 (OETD) and 0.76/0.57 (DECA-UNet).
Whilst fusion often provided reasonable PSMA segmentation, FDG performance degraded and no strategy consistently exceeded the single-tracer baselines.
Under the evaluated setting, tracer-specific models remain the stronger baseline; clinically useful gains from multimodal fusion will likely require architectures that better preserve tracer-specific representations. Our code is available at: \url{https://github.com/JackJ3636/DEEP_PSMA_code}

\end{abstract}

\section{Introduction}
\label{sec:intro}

Accurate assessment of prostate cancer is essential for patient triage and for predicting disease progression and treatment response. Prostate-specific membrane antigen (PSMA) PET/CT frequently alters local radiation planning~\cite{Mena2022TheTreatment}: in an Australian multicentre study, $^{68}$Ga-PSMA PET/CT revealed unsuspected prostate-bed disease in 27\% of patients and changed radiotherapy planning in up to 21\%~\cite{Roach2018TheStudy}. Reliable tumour burden estimation requires accurate lesion delineation, yet manual annotation is time-consuming and subject to intra- and inter-observer variability~\cite{sibille2022wholebodytumorsegmentation18f}, motivating automation~\cite{desouza2022}. Multimodal fusion has shown value across medical applications such as COVID-19 progression prediction~\cite{fang2021}, integrating complementary information much as a clinician combines modalities during clinical decision-making~\cite{krones2026,huang2020fusion}. Existing PET/CT segmentation models perform strongly but are largely limited to a single radiotracer~\cite{alloula2023,dexl2025}. 

In prostate cancer, differences between PSMA and 
fluorodeoxyglucose (FDG) PET tracers, with PSMA reflecting receptor expression and FDG capturing glycolytic activity, further motivate fusion: PSMA excels in typical lesions, whereas FDG may reveal dedifferentiated or PSMA-negative disease \cite{buteau2022therap,chen2025dualtracer}.
Heterogeneous or low PSMA expression is not uncommon with 5-10$\%$ of prostate cancers only showing low-grade PSMA expression \cite{Minner2011HighCancer}, potentially leading to poor uptake in PET imaging \cite{Fendler2023_PSMAGuideline,Tosoian2017_NEPC_DCFPyL}. Meanwhile, FDG PET reflects high-grade, dedifferentiated or PSMA-negative disease, and therefore combining the tracers could therefore capture both metabolic and molecular signatures. However, designing a fusion strategy that integrates complementary features without conflating tracer-specific signatures is non-trivial due to inter-tracer variability, as seen in the AutoPET III Challenge~\cite{rokuss2024}.
This study investigates whether multimodal fusion consistently improves segmentation accuracy over single-tracer baselines.
In this work we (i)~benchmark early- and intermediate-fusion architectures against strong tracer-specific nnU-Net baselines on the public DEEP-PSMA cohort; (ii)~introduce DECA-UNet, a dual-encoder U-Net with zero-initialised channel-wise cross-attention gating between the tracer streams; and (iii)~report a clinically oriented evaluation, with 95$\%$ CIs on mean Dice, showing that under our setup fusion does not consistently outperform single-tracer baselines.
We frame this primarily as an evaluation study and analyse the architectural and data conditions under which fusion is likely to help.

\section{Materials and Methods}
\label{sec:methods}

\subsection{Dataset and Preprocessing}
\label{sec:dataset}

We used the public DEEP-PSMA dataset comprising 100 male patients imaged with paired PSMA ($^{68}$Ga-PSMA-11/$^{18}$F-DCFPyL) and $^{18}$F-FDG PET/CT prior to $^{177}$Lu-PSMA therapy~\cite{jackson2025}.
PSMA and FDG scans were acquired non-simultaneously, so CT inputs were kept tracer-specific (PSMA-CT, FDG-CT). A TotalSegmentator output and a rigid registration parameter file were available to coarsely align the two tracers~\cite{wasserthal2023totalsegmentator}. Each CT was resampled onto its PET grid via the scanner-derived headers (SimpleITK, translation transform, linear interpolation), and for the fusion models FDG PET and CT were resampled into the PSMA coordinate space using the NIfTI affines (54 of 100 cases required resampling). No additional cross-tracer rigid registration was applied beyond the scanner-derived affine alignment.
Preprocessing followed nnU-Net conventions: body cropping; resampling to median spacings ($\approx$3.7--3.8\,$\times$\,3.27\,$\times$\,3.7--3.8\,mm); PET intensities handled as SUV; CT intensities HU-normalised per scan.

A lesion-level analysis showed marked tracer asymmetry: PSMA exhibited more lesions and greater tumour volume than FDG (79.2 vs.\ 45.6 lesions per case; 758.17 vs.\ 238.61\,mL average volume per case), and slightly more spherical lesions (mean sphericity 0.891 vs.\ 0.877). This asymmetry suggests the two segmentation tasks are not equally difficult, which is important context for interpreting the fusion results.
\begin{table}[t]
\centering
\caption{Dice, FP and FN volumes (mL), Surface Dice (SD), SUV\textsubscript{Mean} Ratio (SUV-R), and TTB Volume Ratio (TTB-R) for all models. Mean\,$\pm$\,$\sigma$ reported to two significant figures. $^{\ast}$OEOD predicts a single combined-tracer mask and is evaluated with one combined score; its row is therefore \emph{not} directly comparable to the per-tracer rows and is excluded from the tracer-specific comparisons.}
\label{tab:results}
\setlength{\tabcolsep}{3pt}
\scriptsize
\resizebox{\textwidth}{!}{%
\begin{tabular}{@{}llcccccc@{}}
\toprule
\textbf{Model} & \textbf{Tracer} & \textbf{Dice} & \textbf{FP} & \textbf{FN} & \textbf{SD} & \textbf{SUV-R} & \textbf{TTB-R} \\
\midrule
Baseline FDG     & FDG      & 0.81\,$\pm$\,0.27  & 6.5\,$\pm$\,12   & 18\,$\pm$\,38    & 0.81\,$\pm$\,0.25  & 1.0\,$\pm$\,0.09  & 1.6\,$\pm$\,5.3 \\
Baseline PSMA    & PSMA     & 0.93\,$\pm$\,0.14  & 23\,$\pm$\,52    & 33\,$\pm$\,97    & 0.91\,$\pm$\,0.14  & 1.0\,$\pm$\,0.12  & 1.0\,$\pm$\,0.39 \\
\midrule
OEOD$^{\ast}$ (Early)  & Comb.\   & 0.90\,$\pm$\,0.05  & 0.83\,$\pm$\,1.9 & 3.8\,$\pm$\,5.4  & 0.74\,$\pm$\,0.09  & 0.99\,$\pm$\,0.08 & 0.87\,$\pm$\,0.09 \\
\midrule
OETD (Early)     & PSMA     & 0.69\,$\pm$\,0.32  & 59\,$\pm$\,70    & 120\,$\pm$\,130   & 0.59\,$\pm$\,0.27  & 1.2\,$\pm$\,2.6   & 1.3\,$\pm$\,1.7 \\
                 & FDG      & 0.64\,$\pm$\,0.32  & 26\,$\pm$\,33    & 46\,$\pm$\,56    & 0.55\,$\pm$\,0.27  & 0.53\,$\pm$\,0.63 & 1.2\,$\pm$\,0.88 \\
\midrule
DECA-UNet        & PSMA     & 0.76\,$\pm$\,0.30  & 56\,$\pm$\,67    & 170\,$\pm$\,410  & 0.71\,$\pm$\,0.24  & 0.98\,$\pm$\,0.16 & 0.92\,$\pm$\,0.49 \\
(Intermediate)   & FDG      & 0.57\,$\pm$\,0.31  & 28\,$\pm$\,49    & 56\,$\pm$\,120   & 0.46\,$\pm$\,0.23  & 1.1\,$\pm$\,0.16  & 2.9\,$\pm$\,11 \\
\bottomrule
\end{tabular}%
}
\end{table}
Example coronal slices of segmentation performance are shown for Training Case~96 in Fig.~\ref{fig:qualitative}.

\subsection{Evaluation Metrics}
\label{sec:metrics}

Dice (DSC): volumetric overlap. Surface Dice (SD): surface overlap within a tolerance band, sensitive to boundary quality. FP/FN volume (mL): volume of voxels falsely predicted or missed, i.e.\ ``extra'' or ``missing'' disease burden. SUV\textsubscript{Mean} ratio: predicted-to-true mean SUV inside the mask (near 1.0 indicates matched uptake intensity). TTB volume ratio: predicted-to-true total tumour volume, a therapy-planning biomarker (near 1.0 indicates accurate burden).

\subsection{Baseline Architecture: Tracer-Specific nnU-Net}
\label{sec:baseline}

Two independent 3D full-resolution nnU-Net models were trained as baselines: one for PSMA and one for FDG, each using two input channels (PET and corresponding CT).

\subsection{Fusion Strategies}
\label{sec:fusion}

We investigated two categories of fusion: \textbf{(A)}~Early Fusion and \textbf{(B)}~Intermediate Fusion.

\subsubsection{(A) Early Fusion.}

\noindent\textit{One Encoder, One Decoder (OEOD):} Four input channels (PSMA-PET, FDG-PET, PSMA-CT, FDG-CT) were concatenated and processed by a single 3D nnU-Net encoder--decoder to predict a combined tumour mask with background, tumour, and physiological uptake classes.

\noindent\textit{One Encoder, Two Decoders (OETD):} This approach consists of a shared encoder with tracer-specific segmentation heads. The same four-channel input was encoded once and split into two decoders to produce tracer-specific tumour and physiological-uptake masks (PSMA and FDG heads). Each decoder was supervised with a hybrid Dice\,+\,cross-entropy loss, and the total loss was the average of both heads.

\subsubsection{(B) Intermediate Fusion.}

\noindent\textit{DECA-UNet (Dual-Encoder Cross-Attention U-Net):} This model employs two independent 3D U-Net encoder--decoder pathways, one for PSMA PET/CT and one for FDG PET/CT, each taking a 2-channel input (PET\,+\,CT). The encoders extract hierarchical features at five resolution levels (base channels 32, doubling at each level up to 512).

At encoder levels~3 ($\nicefrac{1}{4}$), 4 ($\nicefrac{1}{8}$), and the bottleneck ($\nicefrac{1}{16}$ resolution), bidirectional cross-attention exchanges information between the PSMA and FDG streams: for each direction the querye is projected from one encoder and the key/value from the other via learnable $1\!\times\!1$ convolutions. Because spatial attention over an $N\!\times\!N$ voxel grid ($N\!=\!D\!\times\!H\!\times\!W$) is prohibitive in 3D, we instead use \emph{channel} attention, forming a $C\!\times\!C$ affinity matrix scaled by $1/\sqrt{C}$, analogous to SE-Net and DANet~\cite{hu2018senet,fu2019}. The attended output is added to the query through a learnable scalar $\gamma$ initialised to zero, so the two encoders behave independently early in training and incorporate cross-tracer information only as $\gamma$ grows:
\begin{equation}
\label{eq:cross_attn}
\mathbf{f}_{\mathrm{out}} = \mathbf{f}_{\mathrm{query}} + \gamma \cdot \mathrm{softmax}\!\left(\frac{\mathbf{Q}\mathbf{K}^{\!\top}}{\sqrt{C}}\right) \mathbf{V}
\end{equation}
where $\mathbf{Q}=W_q(\mathbf{f}_{\mathrm{query}})$ and $\mathbf{K},\mathbf{V}=W_k,W_v(\mathbf{f}_{\mathrm{kv}})\in\mathbb{R}^{C \times N}$ are $1\!\times\!1$ projections and the softmax is applied row-wise over the $C\!\times\!C$ matrix. This zero-initialised gating lets the network learn modality-specific representations before fusing, acting as a form of residual gating.

Each decoder produces 3-class logits (background, tumour, normal uptake).
The two 3-class outputs are combined into a unified 5-class segmentation.
The loss is soft Dice (excluding background), with equal weighting across tracer heads.

\subsection{Implementation Details}
\label{sec:implementation}
All models were implemented in PyTorch and trained on NVIDIA A100 GPUs ($\geq$16\,GB VRAM). For a fair architectural comparison, all fusion variants inherited the self-configured nnU-Net hyperparameters of the tracer-specific baselines: SGD (initial learning rate 0.01, polynomial decay with power 0.9, weight decay $3\!\times\!10^{-5}$), Dice loss, batch size 2, and patch size $112\!\times\!192\!\times\!112$ on 5-fold cross-validation for 200 epochs with standard nnU-Net augmentations (random flips, $90^\circ$ rotations, intensity scaling) and foreground-oversampled patch cropping, following~\cite{isensee2021nnunet}; the self-configuring pipeline removes hyperparameter search as a confounder when comparing architectures. Post-processing was applied to reduce false positives arising from physiological tracer uptake in normal organs.
Model-specific loss functions are described in Sec.\ref{sec:fusion}.

\subsection{Statistical Analysis}
\label{sec:stats}
All models were trained and evaluated under the same 5-fold cross-validation split, so that every patient appears in the same validation fold across architectures.
For each model we report the mean Dice over the full cohort ($N=100$) together with a  $95\%$ confidence interval for the mean.

\section{Results}
\label{sec:results}

\begin{figure}[t]
\centering
\includegraphics[width=\textwidth]{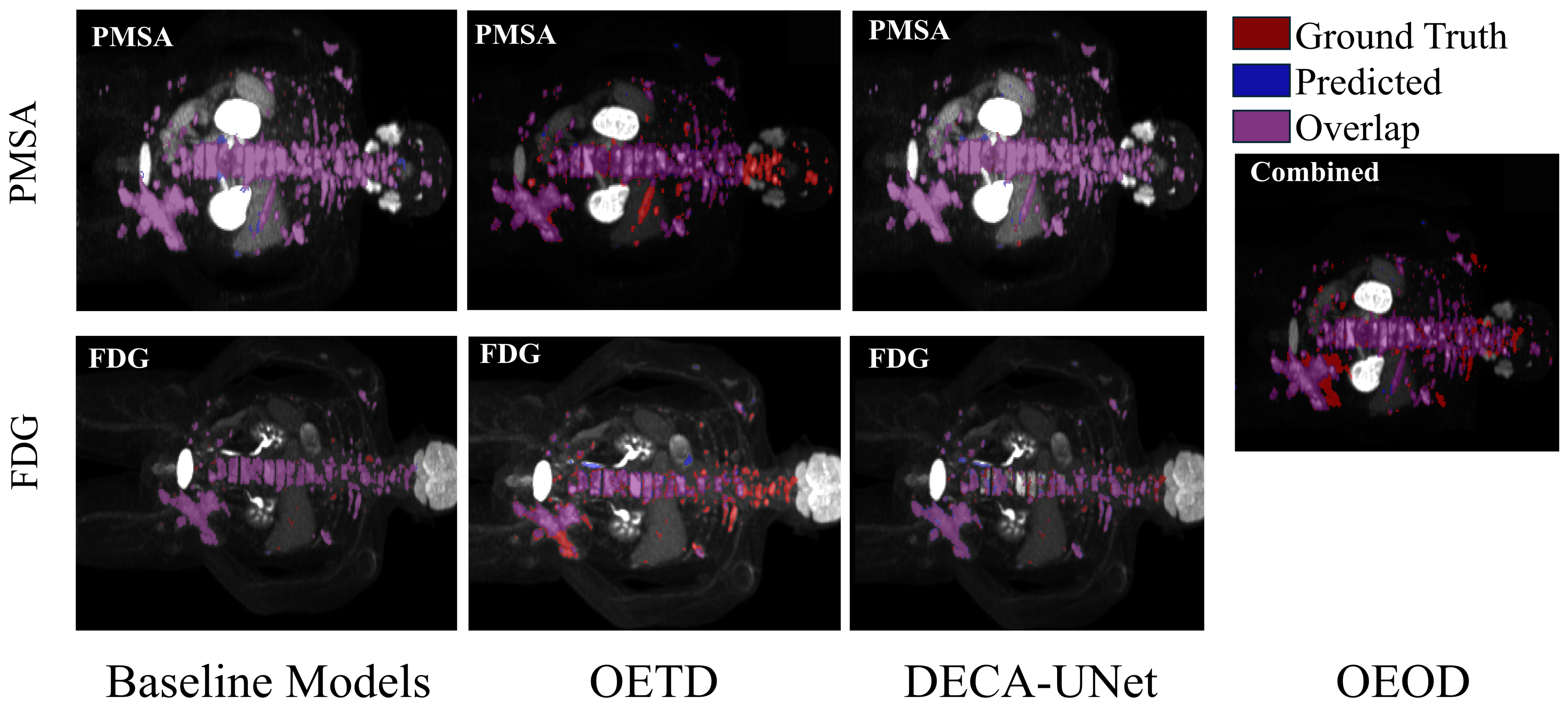}
\caption{Coronal slices of segmentation performance for Baseline models, OETD (early fusion), DECA-UNet (intermediate fusion), and OEOD (early fusion) on Training Case~96 for both FDG and PSMA tracers (combined mapping for OEOD). \textcolor{red}{Red}: ground truth only; \textcolor{blue}{Blue}: prediction only (false positive); \textcolor{purple}{Purple}: overlap between prediction and ground truth (true positive).}
\label{fig:qualitative}
\end{figure}

\subsection{Baseline (Single-Tracer) Performance}

Tracer-specific nnU-Nets were strong and consistent across folds (PSMA Dice=0.93 $\pm$ 0.14, FDG\,=\,0.81\,$\pm$\,0.27; Table~\ref{tab:results}), establishing robust single-tracer benchmarks against which to judge fusion.

\subsection{Early Fusion}

\noindent\textbf{OEOD (combined mask, not directly comparable).}
Early concatenation with a single decoder produced a combined-mask Dice of 0.90\,$\pm$\,0.049.

\noindent\textbf{OETD (dual decoders).}
Sharing an encoder while predicting tracer-specific masks degraded both heads relative to baseline (PSMA/FDG\,=\,0.69/0.64; Table~\ref{tab:results}). Despite the decoder split, the single encoder could not represent both tasks' feature requirements. One possible explanation is representation or gradient conflict arising from sharing an encoder across two tracer-specific tasks.

\subsection{Intermediate Fusion (DECA-UNet)}

The cross-attention model gave a reasonable PSMA Dice (0.76\,$\pm$\,0.30) but a collapse in FDG performance (0.57\,$\pm$\,0.31; Table~\ref{tab:results}).
The high fusion-model variance (${\sim}\pm$\,0.3) reflects both the small cohort (N\,=\,100, 20 test cases per fold) and genuine case-level instability: some cases segmented accurately, others near-zero.

\begin{table}[t]
\centering
\caption{Mean Dice with $95\%$ confidence intervals for the mean ($\bar{x}\pm t_{0.975,99}\,s/\sqrt{N}$, $N=100$), derived from the reported per-case mean and standard deviation. OEOD predicts a single combined-tracer mask and is shown for reference only.}
\label{tab:cis}
\footnotesize
\begin{tabular}{@{}llcc@{}}
\toprule
\textbf{Model} & \textbf{Tracer} & \textbf{Mean Dice} & \textbf{95\% CI} \\
\midrule
Baseline       & PSMA   & 0.93 & [0.90, 0.96] \\
Baseline       & FDG    & 0.81 & [0.76, 0.86] \\
\midrule
OEOD (ref.)    & Comb.\ & 0.90 & [0.89, 0.91] \\
\midrule
OETD           & PSMA   & 0.69 & [0.63, 0.75] \\
OETD           & FDG    & 0.64 & [0.58, 0.70] \\
\midrule
DECA-UNet      & PSMA   & 0.76 & [0.70, 0.82] \\
DECA-UNet      & FDG    & 0.57 & [0.51, 0.63] \\
\bottomrule
\end{tabular}
\end{table}

\noindent\textbf{Uncertainty.}
Table~\ref{tab:cis} reports $95\%$ confidence intervals for each model's mean Dice. The baseline confidence intervals lie above those of both tracer-specific fusion models, providing descriptive evidence of a substantial performance gap. By contrast, the OETD and DECA-UNet intervals overlap substantially on both tracers, so we do not interpret the small differences between the two fusion models as meaningful. The wide fusion intervals (about $\pm0.06$ on the mean, from per-case standard deviations near $0.30$) reflect the limited cohort and the case-level instability noted above.

\subsection{Secondary Metrics and Qualitative Trends}

Secondary metrics tracked the Dice ranking. Surface Dice preserved the same ordering (baselines 0.81/0.91 for FDG/PSMA; OEOD highest among fusion models at 0.74), and OEOD also gave the lowest FP and FN volumes. DECA-UNet produced the most unstable burden estimates, with a high and highly variable FDG TTB ratio (2.9\,$\pm$\,11) and large FP/FN volumes whose wide spreads (\eg\ $\pm$\,410\,mL FN on PSMA) suggest substantial case-level variability.

\begin{figure}[t]
\centering
\includegraphics[width=0.49\textwidth]{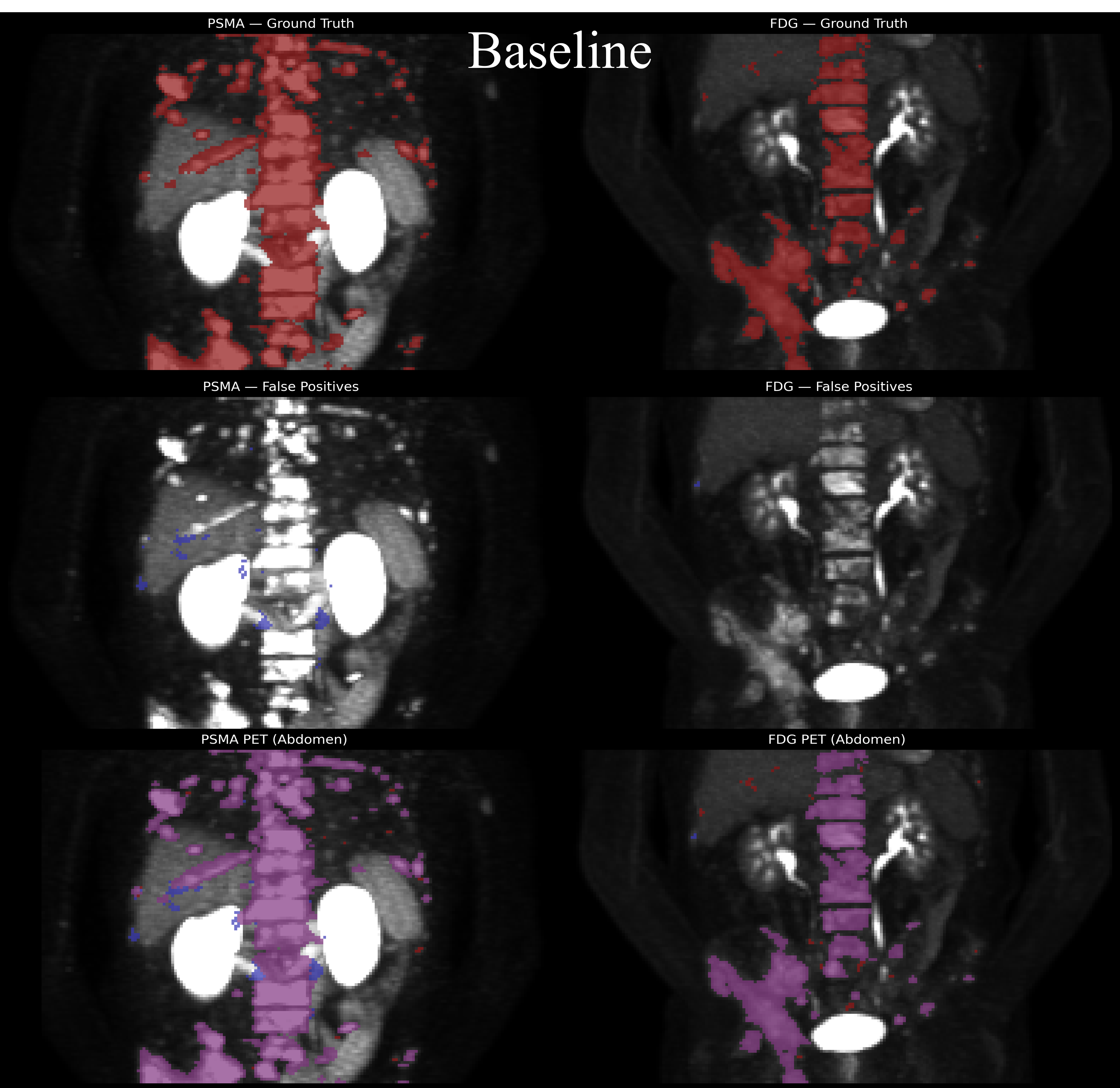}\hfill
\includegraphics[width=0.49\textwidth]{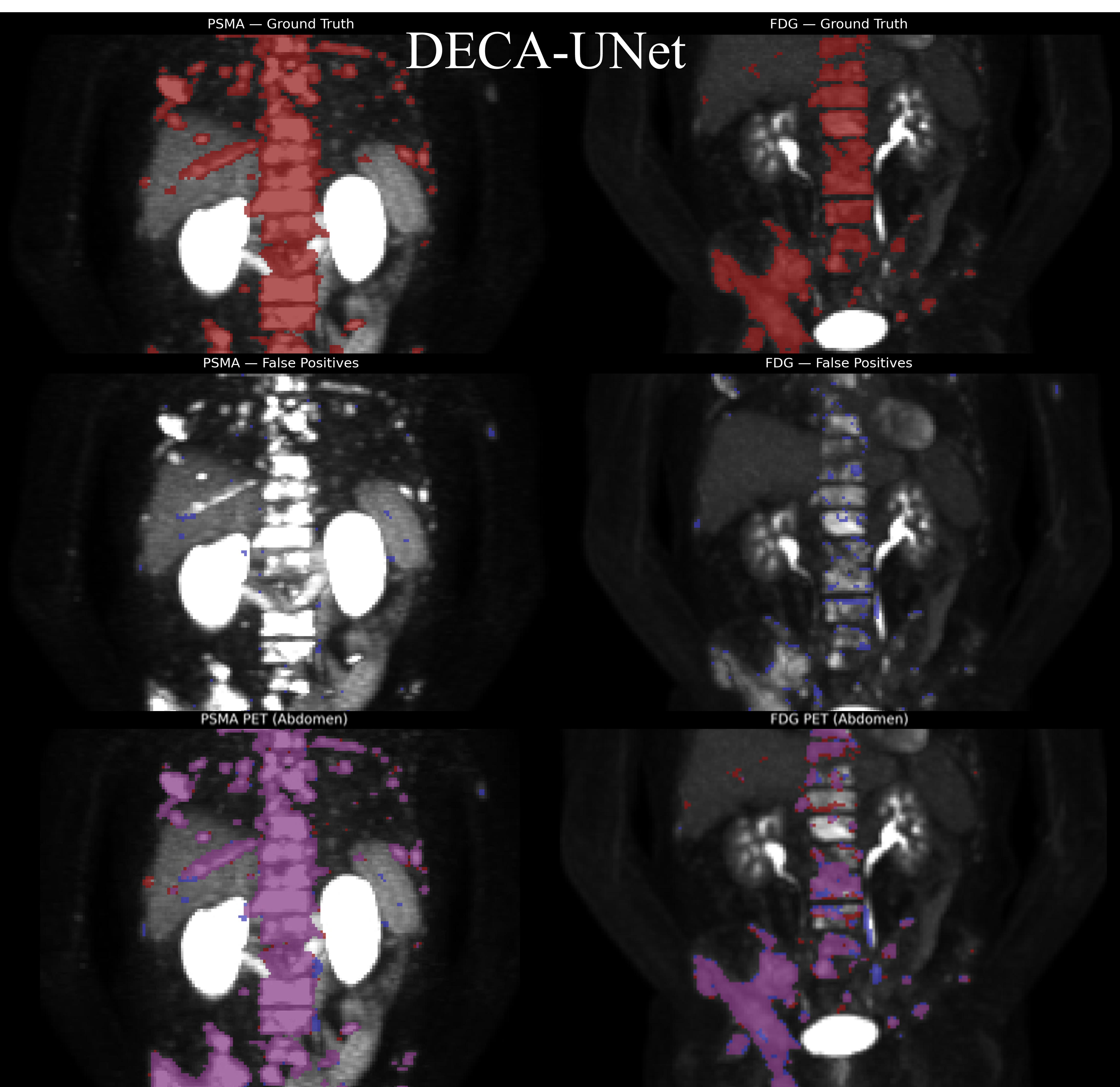}
\caption{Zoomed coronal view comparison of the ground truth, false positives and segmentations for the abdomen region for baseline nnU-Nets (left) and DECA-UNet (intermediate fusion) (right) on Training Case~96. Whilst intermediate fusion did occasionally prevent some false positives in PSMA segmentation (left), this was accompanied by corresponding FDG performance (right) to degrade. \textcolor{red}{Red}: ground truth only; \textcolor{blue}{Blue}: prediction only (false positive); \textcolor{purple}{Purple}: overlap between prediction and ground truth (true positive).}
\label{fig:abdomen}
\end{figure}

The notably low FP volume of OEOD (0.83\,$\pm$\,1.9\,mL) versus the baselines (6.5 and 23\,mL for FDG and PSMA) is largely an artefact of task simplification: predicting a single combined mask absorbs inter-tracer disagreements (which inflate FPs in tracer-specific evaluation) into one label. Conversely, OETD and DECA-UNet show inflated FP volumes (26--59\,mL)consistent with cross-tracer feature mixing allowing avid regions from one tracer to influence the other tracer’s prediction head, notably in the abdomen where physiological uptake patterns diverge (Fig.~\ref{fig:qualitative}, ~\ref{fig:abdomen}).

\section{Discussion}
\label{sec:discussion}

\noindent\textbf{Under what conditions might fusion help?}
We present a number of different suggestions as to why each fusion approach over/underperformed, although attention maps and further lesion-level analyses would be required to confirm these effects. Fusion is most promising when modalities offer complementary evidence for the same focus (\eg, PSMA-avid with modest FDG signal) and when the architecture preserves tracer-specific statistics. OEOD's strong overlap must be read with caution: concatenating tracers reduces the task to detecting \emph{any} lesion. Because tracer provenance is essential for radiotherapy selection, a model that cannot attribute a lesion to its tracer offers limited decision support, so we do not advocate OEOD clinically despite its high Dice.

DECA-UNet applies channel-wise cross-attention across the PSMA and FDG streams; whereas Attention U-Net~\cite{oktay2018} and the gated networks of Schlemper~\etal~\cite{schlemper2019} attend within a single modality, our gating operates \emph{across} tracers. The fact that DECA-UNet was the least degraded fusion variant on PSMA performance suggests attention could potentially guide useful cross-modal transfer, but the FDG drop shows that aggressive feature exchange can help the dominant tracer while hurting the weaker one when biokinetics or lesion phenotypes diverge. Consistent with the lesion analysis, the larger, more numerous PSMA lesions appear to overpower their FDG counterparts, making FDG the harder task.

\noindent\textbf{Why did OETD underperform?} 
A single encoder must model two distinct PET intensity distributions at once, risking representation conflict; despite dual decoders, upstream mixing potentially reduced feature separability and hurt both heads. This cautions against shared encoders unless normalisation and capacity account for tracer heterogeneity.

\noindent\textbf{Limitations and future work.}
Several limitations qualify these findings.
\emph{Cohort size.} With $N\!=\!100$ and 20 test cases per fold, the high variance ($\sim\pm0.30$) limits the conclusions that can be drawn about higher-capacity fusion models and may reflect the difficulty of training such architectures on a limited cohort; the case-level instability we observe (accurate on some patients, near-zero on others) reinforces this. Larger, ideally multi-centre, cohorts are needed to confirm the trends, and the confidence intervals in Sec.~\ref{sec:stats} should be read with this variance in mind. 
\emph{Loss weighting.} Currently, loss is weighted evenly across both heads. However, the disparity in lesion count, volume and sphericity between PSMA and FDG (Sec.~\ref{sec:dataset}), could plausibly cause the gradients from the dominant PSMA to overwhelm those from the sparser FDG task. Thus, exploring uncertainty-based weighting or gradient normalisation approaches would be a promising avenue for future investigation \cite{kendall2018multitasklearningusinguncertainty,chen2018gradnormgradientnormalizationadaptive}. \emph{Registration.} PSMA and FDG were acquired non-simultaneously and aligned only by the scanner-derived affine (Sec~\ref{sec:dataset}). Residual misalignment may contribute to the fusion deficit: DECA-UNet's channel attention pools over space and thus is sensitive to spatial offset, whilst the OEOD and OETD models concatenate tracers at the input and thus assume voxel-level correspondence. This implies that misregistration should penalise early fusion more than intermediate fusion - adding a deformable cross-tracer registration step would assess this directly. \emph{Attribution of the cross-attention effect.} DECA-UNet differs from the single-tracer baseline in two ways at once (dual encoders with multi-task supervision \emph{and} cross-attention gating), so the present design cannot separate the effect of the cross-attention mechanism from that of the dual-encoder multi-task setup. A dual-encoder variant with the gating disabled ($\gamma\!\equiv\!0$) is needed to attribute the effect, and a lesion-level breakdown (by lesion size, FDG-only lesions, and abdominal false positives) would localise where fusion fails; both are priorities for future work. Finally, disagreement-aware attention (suppressing cross-attention where the two tracer representations diverge) could prevent the model from forcing agreement where tracers are expected to differ.

\section{Conclusion}
\label{sec:conclusion}
Multimodal fusion between PSMA and FDG PET/CT did not improve tracer-specific segmentation over strong single-tracer nnU-Net baselines in our experiments. Among the tracer-specific fusion approaches, cross-attention better preserved PSMA performance than FDG performance, revealing a marked asymmetry in fusion performance.
Our results favour tracer-specific models under the evaluated setting, with fusion as a targeted adjunct where complementary tracer patterns are expected and the architecture preserves tracer-specific representations.

\subsubsection*{Compliance with Ethical Standards.}
This research used retrospective, open-access data from the DEEP-PSMA Challenge (MICCAI 2025); ethical approval was not required under the data licence.

\subsubsection*{Acknowledgements}
B.W.P. acknowledges Medical Research Council award (grant no. MR/Y008421/1). 

\bibliographystyle{splncs04}
\bibliography{main}

\end{document}